# Efficient Hate Speech Detection: Evaluating 38 Models from Traditional Methods to Transformers

Mahmoud Abusaqer
Missouri State University
Springfield, Missouri, USA
ma7956s@missouristate.edu

Jamil Saquer
Missouri State University
Springfield, Missouri, USA
jamilsaquer@missouristate.edu

Hazim Shatnawi
The George Washington University
Washington, D.C., USA
hazim.shatnawi@email.gwu.edu

## Abstract

The proliferation of hate speech on social media necessitates automated detection systems that balance accuracy with computational efficiency. This study evaluates 38 model configurations in detecting hate speech across datasets ranging from 6.5K to 451K samples. We analyze transformer architectures (e.g., BERT, RoBERTa, DistilBERT), deep neural networks (e.g., CNN, LSTM, GRU, Hierarchical Attention Networks), and traditional machine learning methods (e.g., SVM, CatBoost, Random Forest).

Our results show that transformers, particularly RoBERTa, consistently achieve superior performance with accuracy and F1-scores exceeding 90%. Among deep learning approaches, Hierarchical Attention Networks yield the best results, while traditional methods like CatBoost and SVM remain competitive, achieving F1-scores above 88% with significantly lower computational costs.

Additionally, our analysis highlights the importance of dataset characteristics, with balanced, moderately sized unprocessed datasets outperforming larger, preprocessed datasets. These findings offer valuable insights for developing efficient and effective hate speech detection systems.

## 1 Introduction

The rapid growth of social media has fundamentally transformed online discourse, bringing unprecedented connectivity but also new challenges in content moderation. As platforms struggle to balance free expression with user safety, the detection and mitigation of hate speech has become increasingly critical [31]. While manual moderation remains valuable, the scale of online communication necessitates robust automated detection methods [40].

Early approaches to hate speech detection relied primarily on traditional machine learning (ML) techniques such as Support Vector Machines (SVM), Naive Bayes (NB), and Random Forests (RF) [40, 44]. However, these methods often struggle with context-dependent expressions and evolving linguistic patterns. Recent advances in deep learning (DL) transformer architectures have shown promise in capturing semantic nuances [30], yet questions remain about their practical deployment at scale. Three key questions emerge from current research:

(1) How do transformer architectures compare with traditional DL and ML approaches when evaluated systematically across different hate speech datasets? While recent work suggests the potential of BERT-based models [18, 39, 43], comprehensive comparisons remain limited.
(2) What is the relationship between preprocessing, dataset composition, and model performance? Prior studies report conflicting results on preprocessing effectiveness [1, 45], particularly for transformer models.
(3) How do various DL architectures perform when scaled to production-level datasets? Existing evaluations typically focus on smaller datasets [18], leaving questions about real-world applicability.

## 2 Research Contributions

This study advances the field in three key directions. First, we present a comprehensive evaluation of hate speech detection approaches across multiple scales. We combine three widely used, relatively small hate speech datasets into a mid-size dataset of 48,049 samples: Davidson et al.'s [9] hate speech corpus comprising 24,783 tweets, the Impermium dataset [21] with 6,594 social media comments, and the Waseem and Hovy's Twitter dataset [48] containing 16,907 tweets. Our experimental results on this combined dataset establish significant performance benchmarks, with RoBERTa achieving 91.48% accuracy and 92.62% F1-score.

For scaled evaluation, we employ multiple dataset configurations, shown in Table 1, following Mody et al. [34]: a large unbalanced preprocessed dataset of 451,709 samples, maintaining natural class distributions found in social media environments; a large balanced preprocessed dataset of 160,500 samples; and a balanced raw dataset of 267,388 samples, which enables evaluation of model performance on unmodified content while controlling for class imbalance. We use a total of seven different size datasets ranging from small (6,594 samples) to large (451,709 samples) for our experiments.

Second, we analyze the impact of dataset characteristics through systematic experimentation, building upon the methodologies of Ross et al. [38] and Wiegand et al. [49]. Our findings challenge common assumptions about dataset size, demonstrating that raw balanced datasets of moderate size (267K samples) achieve higher performance of 91.89% accuracy and F1-score with RoBERTa compared to both a larger processed unbalanced 451K dataset with 90.72% accuracy and 90.54% F1-score and a combined mid-size dataset of 48,049 samples, where RoBERTa achieves 91.48% accuracy and 92.62% F1-score.

The effect of preprocessing varies by architecture: RoBERTa model performance drops from 91.48% to 89.40% accuracy. Most DL models show performance reduction regardless of embedding choice. For example, HAN with GloVe decreases from 87.18% to 86.90% accuracy, LSTM with Word2Vec drops from 86.17% to 82.77% accuracy, while BiGRU with GloVe increases from 86.70% to 86.77% accuracy. Traditional ML models maintain relatively stable performance with best results shifting from CatBoost at 88.60% to SVM RBF at 88.56% accuracy.

Third, we demonstrate the continuing relevance of traditional approaches using implementations described by Liaw et al. [28], with CatBoost achieving 88.60% accuracy and 89.60% F1-score, and SVM reaching 88.50% accuracy and 89.50% F1-score. These methods offer advantages in computational efficiency and interpretability, suggesting their viability for resource-constrained deployments.

We perform extensive experiments employing 38 different traditional ML, DL, and transformer models over seven datasets. Our findings have significant implications for practical implementations. While transformer models establish new performance benchmarks, our results indicate that carefully tuned traditional approaches often provide better efficiency-performance trade-offs.

## 3 Literature Review

Research in automated hate speech detection spans multiple disciplines, encompassing both social science and computer science perspectives [12]. Terms like "cyberbullying" and "online harassment" are closely associated with online hate speech detection efforts, which aim to identify and prevent hate crimes [51].

Early studies relied on feature engineering and traditional ML approaches such as NB and SVM [40, 44]. Waseem and Hovy established foundational datasets demonstrating the effectiveness of character n-grams and demographic features [48]. NLP techniques that identify and categorize abusive language targeting certain characteristics, and utilizing supervised learning models like Logistic Regression (LR), SVM, Random Forests, and DL approaches have been used [4, 42]. ElSherief et al. [11] conducted comprehensive analysis of hate speech patterns using linguistic and psycholinguistic approaches. Their work incorporated NLP techniques including part-of-speech tagging and sentiment analysis, establishing the effectiveness of multi-faceted feature extraction. The work in [43] achieved best result of 0.756 F1-score on SVM among several traditional ML models, which we surpassed in our experiments by more than 10% on all traditional ML methods as shown in Table 4.

Deep learning architectures have shown superior capabilities in capturing language complexities. Liu [29] demonstrated CNN and LSTM effectiveness with F1-scores of 0.800 and 0.778, respectively, on the Waseem and Hovy dataset. Shawkat et al. showed BERT outperformed both traditional ML and other deep learning approaches with F1-scores above 0.88 across multiple datasets [43]. That research did not evaluate the performance of other transformers against BERT. Saleh et al. achieved state-of-the-art results using BERT with specialized word embeddings, demonstrating F1-score of 0.96 on their large balanced dataset of 1,048,563 tweets [39]. The authors have not made their dataset public and they did not compare BERT with other transformers. These results underscore the superiority of transformer architectures like BERT in capturing contextual relationships, motivating further exploration of other transformers such as RoBERTa, XLM-RoBERTa, and DistilBERT, which we undertake as part of the work in this paper.

Multi-modal approaches incorporating video, audio, and textual data have emerged as promising directions. Boishakhi et al. demonstrated the versatility of machine learning in identifying hate speech across different media formats [2]. However, challenges persist, including language ambiguity, lack of standardized definitions, and evolving hate speech patterns. Dataset characteristics significantly impact model performance, as shown by Sintaha and Mostakim's evaluation across different configurations [46].

Recent work addresses implementation challenges. Agrawal and Awekar [1] explored platform-specific characteristics in cyberbullying detection, while Khan and Phillips [23] developed language-agnostic models for detecting Islamophobic content. The relationship between online hate speech and real-world consequences has gained attention, with studies by Giumetti and Kowalski demonstrating significant impacts on mental well-being [16].

Despite advances in hate speech detection, key challenges persist: the need for comprehensive architectural evaluation and production scalability; platform-specific variations and evolving language patterns requiring robust preprocessing and language-agnostic approaches; and fundamental issues of linguistic ambiguity, unstandardized definitions, and constantly evolving hate speech forms necessitating continuous algorithm adaptation. These interconnected challenges demand ongoing research to develop more resilient detection systems.

## 4 Datasets

In this section, we present the datasets used in our experiments. First, We implemented experiments using three well established, relatively small datasets following the approach in Shawkat et al. [43]. Then, we combined these datasets into a mid-size dataset for more experiments. For further analysis, we used three additional large scale datasets.

### 4.1 Small Datasets

Our initial experiments utilized three datasets that are widely used in hate speech detection, which we will refer to as small datasets. Table 1 shows summaries and class distributions for all datasets used in our research. In the acronyms used, L means Large, B Balanced, U Unbalanced, P Processed, R Raw/Unprocessed, S Small, and M Mid-size.

**Table 1: Datasets Used with Hate Speech Class Distribution**

| Dataset | Hateful Tweets | Non-hateful Tweets | Total Tweets |
|---|---|---|---|
| I. Waseem & Hovy (SU) | 5,399 (32.38%) | 11,273 (67.62%) | 16,672 |
| II. Impermium (SU) | 1,742 (26.42%) | 4,852 (73.58%) | 6,594 |
| III. Davidson (SU) | 20,620 (83.20%) | 4,163 (16.80%) | 24,783 |
| IV. Combined Mid (M Roughly Balanced) | 27,761 (57.87%) | 20,288 (42.13%) | 48,049 |
| V. Balanced Preprocessed (LBP) | 80,250 (50%) | 80,250 (50%) | 160,500 |
| VI. Balanced Raw (LBR) | 133,694 (50%) | 133,694 (50%) | 267,388 |
| VII. Unbalanced Preprocessed (LUP) | 80,250 (17.8%) | 371,452 (82.2%) | 451,709 |

a) Dataset I was created by Waseem and Hovy [48] and contains tweets that were manually labeled for hate speech by human annotators. The dataset exhibits the following distribution: 1,976 tweets labeled as racism (11.85%), 3,423 as sexism (20.53%), and 11,273 as neither racist nor sexist (67.62%). Following previous work [11, 43], we combined the racism and sexism categories into a single hateful class, resulting in 5,399 hateful tweets (32.38%) and 11,273 non-hateful tweets (67.62%).

b) Dataset II, released by Impermium [21], comprises 6,594 social media comments from multiple platforms including Facebook/Meta, Twitter/X, and Reddit. The data is structured with 3,947 training instances (1,049 insulting, 2,898 non-insulting) and 2,647 testing instances (693 insulting, 1,954 non-insulting). After combining the training and testing sets, this results in 1,742 hateful (26.42%) and 4,852 non-hateful (73.58%) posts.

c) Dataset III, developed by Davidson et al. [9], contains 24,783 tweets with a three-way classification scheme: 1,430 tweets labeled as hate speech (5.77%), 19,190 as offensive language (77.43%), and 4,163 as neither (16.80%). Following recent approaches [39, 43], we combined hate speech and offensive language categories into a single hateful class, resulting in 20,620 hateful tweets (83.20%) and 4,163 non-hateful tweets (16.80%).

## 4.2 Combined Mid-Size Dataset

Building upon recent work that demonstrates the benefits of larger training sets for text classification and hate speech detection [18, 39], we merged the three small datasets into a single combined dataset, which we refer to as "Combined Mid." This combined dataset contains 48,049 total entries. We carefully handled label harmonization across the different source datasets, ensuring consistent binary classification (hateful vs. non-hateful) across all entries, following best practices established in recent literature [38, 49].

## 4.3 Large-Scale Dataset Configurations

Our experimental evaluation extends to three larger-scale datasets:

*4.3.1 Unbalanced Preprocessed Dataset (451K Samples).* We used the dataset created by Mody et al. [34], which contains 451,709 samples in total. The dataset exhibits the following distribution: 371,452 entries labeled as non-hateful (82.2%) and 80,250 as hateful (17.8%). This configuration maintains the unbalanced class distribution found on social media environments. The dataset was curated by combining data from 18 sources such as Kaggle and GitHub. The dataset underwent comprehensive preprocessing such as removing duplicates and special characters, converting emojis to text, and expanding contractions.

*4.3.2 Balanced Preprocessed Dataset (160K Samples).* This dataset was created by balancing the unbalanced preprocessed dataset (451K samples) from Mody et al. [34], resulting in a total of 160,500 samples. Balancing was done by undersampling the majority class. We maintained equal class distribution with 80,250 samples per class while preserving all preprocessing steps from the original dataset. This balanced configuration enables unbiased evaluation of model performance while controlling for class imbalance effects.

*4.3.3 Balanced Raw Dataset (267K Samples).* For this dataset, we utilized the raw dataset from Mody et al. [34]. which is provided as a separate file and contains over 800K entries in multiple languages. We filtered the dataset to include only English-language content and implemented class balancing, resulting in 267,388 total entries. The balanced distribution consists of 133,694 hateful (50%) and 133,694 non-hateful (50%) samples. By maintaining the raw text characteristics while controlling for class balance, this dataset allows us to evaluate model performance on unmodified content without bias from class distribution.

These dataset configurations complement the small and mid-size datasets, enabling comprehensive evaluation across varying scales and conditions. Each configuration addresses specific aspects of model evaluation, from unbalanced class distribution to controlled balanced scenarios.

# 5 Data Preprocessing

In this section, we detail our preprocessing pipeline, adapted from Mody et al. [34], which we apply to the "Combined Mid." and other datasets as needed. The preprocessing consists of the following tasks:

## 5.1 Text Cleaning and Normalization

- Structural Cleaning:
  - Remove URLs using regular expressions (http, www, https)
  - Remove user mentions (@username)
  - Convert all text to lowercase
  - Remove HTML tags and artifacts (e.g., encoded entities &, “, <, >, broken tags <>, </, markup fragments <div, /span>)
- Content Normalization:
  - Expand contractions using the contractions library (e.g., "won't" → "will not")
  - Convert emojis and emoticons to text

- Handle dates and times with custom formatting
- Convert numerical values to words using the inflect library

## 5.2 Linguistic Processing

- Stop Word Removal:
  - Implement a custom stop words list while preserving context-important words
  - Apply word-level filtering to remove common English stop words
- Text Standardization:
  - Remove special characters such as (\-;%&=*/%#$@) while preserving essential punctuation
  - Maintain only alphabetic characters
  - Remove extra whitespace and standardize spacing

Our preprocessing pipeline focuses on fundamental text cleaning and normalization steps to optimize our machine learning and deep learning models' performance.

# 6 Methodology

Our research presents a comprehensive evaluation of ML and DL approaches for hate speech detection, encompassing multiple model architectures, embedding techniques, and training methodologies. This section details our experimental framework, providing in-depth analysis of each component and implementation decision.

For this study, we implemented diverse model architectures spanning traditional ML, deep neural networks (NN), and state-of-the-art transformer models. All models underwent rigorous evaluation using 5-fold cross-validation. For DL and transformer models, we employed early stopping mechanisms implemented after 3 epochs without improvement to ensure optimal convergence. Our implementation leverages both PyTorch and TensorFlow frameworks.

The following subsections detail each model category, their architectural choices, and implementation specifications:

*6.0.1 Text Representation.*

- TF-IDF: Weights terms by their frequency in a document and inverse frequency across documents [15]. This is usually used with traditional ML approaches. Since TF-IDF performed better than n-grams (sequences of n consecutive words) in [43], we used TF-IDF in our experiments with traditional ML models.
- GloVe: It generates dense vector representation of words that capture their semantic and syntactic relationships. GloVe uses a global matrix factorization approach to directly optimize word vectors based on their co-occurrence counts [35].
- Word2Vec: It generates word embeddings by using skip-gram and continuous bag-of-words architectures [33]. Both approaches create dense vector representations where similar words cluster together and semantic relationships are preserved through vector operations.

*6.0.2 Transformer Models.* Implemented using PyTorch and the HuggingFace Transformers library with Ray Tune [28] employed for hyperparameter optimization:

- BERT: The foundational bidirectional transformer model that uses masked language modeling for pret-raining [10]. It follows a bidirectional approach to learn contextual relations between words in a text that understands a word based on its surroundings from both directions.
- RoBERTa: It enhances BERT's performance by removing the next sentence prediction objective and implementing dynamic masking during training. It employs larger batch sizes, higher learning rates, and more extensive training data [30]. These optimizations typically result in improved performance across various NLP tasks.
- DistilBERT: A lighter version of BERT created through knowledge distillation. It reduces the model size by 40% through reducing layers, while maintaining 97% of BERT's language understanding capabilities [41]. The architecture removes the token-type embeddings and pooler while keeping the general transformer architecture. This optimization makes DistilBERT faster and more resource-efficient while still delivering strong performance on NLP tasks.
- ALBERT (A lite BERT): It reduces BERT's memory usage and training time through two key techniques: cross-layer parameter sharing, where parameters are shared across transformer layers, and factorized embedding parameterization, which decomposes the large vocabulary embedding matrix into smaller matrices. These optimizations significantly reduce parameters without substantially impacting model performance. ALBERT also replaces BERT's Next Sentence Prediction with Sentence Order Prediction for better inter-sentence coherence learning [26].
- XLM-RoBERTa: A multilingual version of RoBERTa trained on 100 languages using a large-scale dataset of 2.5 TB of cleaned "Common Crawl" data. It learns cross-lingual representations through masked language modeling without requiring explicit language embeddings or translation pairs [7].

*6.0.3 Deep Neural Architectures.* We implemented a comprehensive suite of deep learning architectures using TensorFlow, each designed to capture different aspects of textual information for hate speech detection. Our implementation incorporates both GloVe (100-dimensional) and Word2Vec (300-dimensional) embeddings to provide rich semantic representations. All architectures were trained using the Adam optimizer with binary cross-entropy loss and implemented with early stopping (patience of 3) to prevent overfitting.

- Convolutional Neural Network (CNN): Employs convolutional operations to extract local text features [27]. Our architecture consists of an embedding layer followed by a Conv1D layer (128 filters, kernel size 5) with ReLU activation, global max pooling, and a dense layer (64 units) with dropout (0.5), followed by a dense output layer with sigmoid activation for classification.
- TextCNN: Extends the basic CNN by utilizing multiple parallel convolutional layers with different filter sizes (3,4,5) to capture varied n-gram features [24]. These parallel outputs are concatenated and processed through a dense layer with dropout.
- Long Short-Term Memory (LSTM): Specifically designed to handle long-term dependencies in text sequences [19]. Our implementation uses two stacked LSTM layers (64 and 32 units) with a final dense layer.

- BiLSTM: Enhances the LSTM by processing sequences bidirectionally [17], allowing the model to capture context from both past and future tokens. Uses the same layer configuration as LSTM but with bidirectional processing.
- Gated Recurrent Unit (GRU): Provides a more computationally efficient alternative to LSTM [6] while maintaining comparable performance. Implements two GRU layers followed by a dense layer with dropout.
- CNN-GRU and CNN-LSTM: Hybrid architectures combining CNN's local feature extraction capabilities with sequential processing [47, 53]. Both implement a Conv1D layer (128 filters) followed by their respective sequential processors (GRU or LSTM).
- Recurrent CNN (RCNN): Integrates recurrent connections into the CNN architecture [25] to capture both local and sequential patterns. Combines a recurrent layer with convolutional operations and global pooling.
- Hierarchical Attention Networks (HAN): Leverages the natural structure of text, where words form sentences and sentences form documents [52]. The architecture implements two levels of attention mechanisms: word-level attention (64 units) identifies important words in sentences, while sentence-level attention (32 units) identifies important sentences in documents. The model processes text through an embedding layer using pre-trained word embeddings, followed by a bidirectional LSTM returning sequences. The attention-weighted representations are aggregated through global max pooling, followed by dense layers for final classification.
- Graph Neural Network (GNN): Represents text as a graph structure [50], enabling the capture of complex relationships between words. Our model architecture employs two sequential graph attention layers with 64 units in the first layer followed by 32 units in the second layer, combined with convolutional processing.

Each architecture was implemented with careful consideration of modern deep learning practices, including proper regularization through dropout and batch normalization where appropriate. The specific configurations were chosen based on experimentation and established best practices in the literature.

*6.0.4 Traditional Machine Learning.* Implemented using scikit-learn with TF-IDF features:

- CatBoost: An advanced implementation of gradient boosting that handles categorical features through ordered boosting. The algorithm employs a symmetric tree structure and implements ordering principle to minimize prediction shifts, featuring novel techniques for processing categorical variables and automatic handling of missing values [36].
- SVM (RBF kernel): A Support Vector Machine implementation utilizing the Radial Basis Function kernel for non-linear classification. The algorithm projects input data into a higher-dimensional space where it becomes linearly separable, with the RBF kernel enabling efficient non-linear classification through the kernel trick [8].
- LightGBM: A gradient boosting framework that employs histogram-based algorithms for efficient training. The method implements exclusive feature bundling (EFB) to handle high-dimensional data and Gradient-based One-Side Sampling (GOSS) to maintain accuracy while reducing data size. These optimizations enable faster training while preserving model performance [22].
- XGBoost: A scalable tree boosting system implementing a sparsity-aware algorithm for sparse data handling and weighted quantile sketch for approximate tree learning. The framework adds a regularization term to the objective function to control model complexity and implements an efficient block structure for parallel tree construction [5].
- Logistic Regression: A classification algorithm that extends linear regression by applying a logistic function to constrain outputs between 0 and 1. This transformation enables the model to predict class probabilities rather than continuous values, making it suitable for binary classification tasks [20].
- Random Forest: An ensemble learning method that combines multiple decision trees through bootstrap aggregation (bagging). Each tree is trained on a random subset of features and data points, with final predictions made through majority voting [3].
- AdaBoost: An adaptive boosting algorithm that iteratively focuses on misclassified examples by adjusting sample weights. The method implements a sequential ensemble where each weak learner compensates for the mistakes of previous models, with prediction weights determined by individual model accuracy [13].
- Decision Tree: A supervised learning algorithm that creates a hierarchical, tree-like structure for classification. At each internal node, the algorithm selects the most informative attribute using metrics like information gain to split the data. The resulting tree structure makes decisions by following paths from root to leaf nodes, providing interpretable classification rules for new data [37].
- Gradient Boosting: An ensemble technique that produces a strong learner by iteratively adding weak learners to minimize a differentiable loss function. The method implements second-order approximation for faster convergence and employs subsampling strategies for improved generalization [14].
- Naive Bayes (NB): A probabilistic classifier based on Bayes' theorem that calculates class probabilities for new data using prior probabilities learned from training data [32]. While it makes a "naive" assumption of feature independence, it remains effective for text classification tasks. We employed two variants: Multinomial NB, which considers term frequencies, and Bernoulli NB, which focuses on binary word occurrences, as both are particularly well-suited for text analysis.

## 7 Results and Discussion

We trained our models on a system with two NVIDIA Ampere A100 GPUs (6912 CUDA cores, 40GB RAM each) and 512 GB RAM. Our experimental results demonstrate the effectiveness of various approaches to hate speech detection across different model architectures and dataset configurations. We present a detailed analysis of model performance, examining the impact of architectural choices and data characteristics.

**Table 2: Results of Transformer Models on "Combined Mid." Dataset**

| Model | Accuracy | F1 | Precision | Recall |
|---|---|---|---|---|
| **Raw Data** | | | | |
| RoBERTa | **91.48** | **92.62** | 92.73 | 92.50 |
| BERT | 91.40 | 92.55 | 92.68 | 92.41 |
| DistilBERT | 91.38 | 92.52 | 92.72 | 92.32 |
| ALBERT | 90.92 | 92.01 | **93.64** | 90.43 |
| XLM-RoBERTa | 90.61 | 91.97 | 90.84 | **93.13** |
| **Preprocessed Data** | | | | |
| BERT | **89.74** | **90.85** | **93.64** | 88.23 |
| RoBERTa | 89.40 | 90.79 | 92.32 | **90.77** |
| XLM-RoBERTa | 89.07 | 90.36 | 92.64 | 88.09 |
| DistilBERT | 89.27 | 90.65 | 91.24 | 90.07 |
| ALBERT | 88.75 | 90.21 | 90.49 | 89.94 |

## 7.1 Small and "Combined Mid." Datasets Performance Analysis

*7.1.1 Transformer Models Performance.* We developed these models using PyTorch and the HuggingFace Transformers library. Table 2 illustrates that transformer architectures consistently achieve good results on the "Combined Mid." dataset. In raw data experiments, RoBERTa demonstrates the highest performance with 92.62% F1-score and 91.48% accuracy, followed by BERT 92.55% F1-score, 91.40% accuracy and DistilBERT 92.52% F1-score, 91.38% accuracy. With preprocessed data, BERT achieves 90.85% F1-score and 89.74% accuracy, while RoBERTa maintains strong performance with 90.79% F1-score and 89.40% accuracy. The performance of transformers on the raw data is better than their performance on processed data by about 2%.

The performance of transformer models on the small datasets (Dataset 1, 2, and 3) can be seen in Table 8 with RoBERTa achieving the best results. Compared to the work in [43], where they evaluated BERT, our results show better results, with all transformer models achieving F1-score above 97% on Dataset 3, an improvement of about 7%. Our results on Datasets 1 and 2 are only slightly higher than those in [43].

*7.1.2 Deep Neural Architecture Performance.* Table 3 shows the results of the deep neural architectures on the "Mid Combined" dataset. Hierarchical attention networks (HAN) with GloVe embeddings achieve the best results among deep neural architectures, obtaining 88.64% F1-score and 87.18% accuracy. BiGRU and BiLSTM follow with F1-scores of 88.17% and 87.88% respectively. Models utilizing Word2Vec embeddings show slightly lower but consistent performance, with HAN achieving 88.00% F1-score and 86.38% accuracy. Results of the traditional DL models on the other datasets are shown in Appendix A.

*7.1.3 Traditional ML Models Performance.* Table 4 shows the results of the traditional ML models on the "Mid Combined" dataset. Table 4 demonstrates that traditional approaches maintain competitive performance. CatBoost leads with 89.60% F1-score and 88.60% accuracy in raw data experiments followed closely by SVM with RBF kernel with 89.50% F1-score and 88.50% accuracy. In preprocessed data configurations, SVM RBF leads with 89.58% F1-score and 88.56% accuracy followed closely by CatBoost. These results show the competitiveness of traditional ML methods such as SVM and CatBoost, which are higher than many of the DL methods shown in Table 3, and slightly lower then transformer results. Results of the traditional ML models on the other datasets are shown in Appendix A.

## 7.2 Impact of Dataset Characteristics

Our analysis reveals several important trends regarding dataset characteristics and model performance. The combined dataset generally yielded higher performance across all models compared to individual datasets as shown in Table 8, suggesting that diverse training data improves model robustness. This effect was particularly pronounced for transformer-based models, which showed performance improvements when trained on the combined dataset.

## 7.3 Dataset Scale Analysis

The performance across different dataset scales reveals important patterns in model behavior:

*7.3.1 Large-Scale Implementation (451K Samples).* Results on the unbalanced preprocessed dataset demonstrate strong performance at scale, as shown in Table 5. RoBERTa achieves 90.72% accuracy and 90.54% F1-score, while BERT maintains comparable performance with 90.71% accuracy and 90.57% F1-score.

*7.3.2 Balanced Raw Dataset (267K Samples).* As presented in Table 6, model performance peaks with the balanced raw dataset. RoBERTa achieves the highest results with 91.89% in both accuracy and F1-score, while BERT follows with 91.20% in both metrics. This configuration demonstrates the benefits of balanced class distribution with sufficient data volume.

*7.3.3 Balanced Preprocessed Dataset (160K Samples).* Results from Table 7 show that on the balanced preprocessed dataset, RoBERTa leads with 86.46% accuracy and 86.45% F1-score, followed by BERT with 86.32% in both metrics. The lower performance compared to the raw balanced dataset suggests that preprocessing may have a negative impact on model effectiveness.

Analysis across different dataset scales reveals compelling patterns in model behavior and dataset characteristics. The balanced raw dataset (267K samples) demonstrates superior performance with RoBERTa achieving 91.89% in both accuracy and F1-score, followed closely by BERT at 91.20%. However, when examining the balanced preprocessed dataset (160K samples), we observe a significant performance degradation, with RoBERTa achieving only 86.46% accuracy and 86.45% F1-score despite the maintained class balance. Interestingly, the unbalanced preprocessed dataset (451K samples) shows intermediate performance at 90.72% accuracy and 90.54% F1-score, suggesting that preprocessing impact may outweigh both dataset size and class distribution effects. The substantial performance gap between raw and preprocessed configurations indicates that preserving original text characteristics could be more critical for model effectiveness than conventional dataset design considerations, particularly when scaling to larger implementations. This finding aligns with observations from other research. For instance, Siino et al. [45] questioned whether "text preprocessing is still worth the time" and found that preprocessing can negatively impact transformer performance. Similarly, Agrawal and Awekar [1] reported conflicting results on preprocessing effectiveness, particularly for transformer models. Transformer models are pre-trained on natural language in its original form, allowing them to capture nuanced contextual relationships and linguistic patterns. Preprocessing the data may inadvertently eliminate valuable contextual signals that transformers can effectively utilize.

Table 3: Results of Deep Neural Architectures on "Combined Mid." Dataset

| Model | Raw | | | | Preprocessed | | | |
|---|---|---|---|---|---|---|---|---|
| | Accuracy | F1 | Precision | Recall | Accuracy | F1 | Precision | Recall |
| **Models with GloVe Embedding** | | | | | | | | |
| HAN | **87.18** | **88.64** | **90.85** | 86.60 | **86.90** | **88.38** | 90.70 | **86.99** |
| BiGRU | 86.70 | 88.17 | 90.83 | 85.74 | 86.77 | 88.30 | 90.31 | 86.42 |
| BiLSTM | 86.34 | 87.88 | 90.36 | 85.77 | 86.65 | 88.29 | 89.72 | 86.99 |
| LSTM | 86.44 | 87.86 | 89.86 | 85.94 | 84.29 | 85.58 | 89.89 | 81.66 |
| GRU | 86.40 | 87.78 | 90.11 | 85.56 | 84.65 | 86.01 | 89.64 | 82.66 |
| TextCNN | 86.11 | 87.60 | 90.47 | 85.01 | 85.61 | 87.10 | 90.41 | 84.11 |
| CNN | 85.82 | 87.46 | 88.29 | **86.64** | 82.87 | 84.66 | 86.61 | 82.79 |
| CNN-GRU | 85.72 | 87.25 | 90.23 | 84.58 | 85.61 | 87.02 | **90.84** | 83.59 |
| CNN-LSTM | 85.53 | 87.22 | 89.19 | 85.50 | 85.32 | 86.74 | 90.71 | 83.27 |
| RCNN | 85.70 | 87.17 | 90.57 | 84.32 | 85.56 | 87.08 | 90.22 | 84.30 |
| GNN | 82.38 | 84.55 | 84.62 | 84.48 | 78.23 | 79.52 | 85.86 | 74.06 |
| **Models with Word2Vec Embedding** | | | | | | | | |
| HAN | **86.38** | **88.00** | 89.77 | **86.82** | **86.27** | **87.85** | 89.94 | **85.93** |
| LSTM | 86.17 | 87.75 | 88.74 | 86.78 | 82.77 | 84.27 | 87.98 | 80.86 |
| BiLSTM | 86.16 | 87.71 | 90.24 | 85.43 | 85.91 | 87.49 | 89.99 | 85.19 |
| TextCNN | 86.21 | 87.66 | **90.69** | 85.03 | 85.70 | 87.18 | **90.43** | 84.31 |
| GRU | 85.96 | 87.59 | 88.38 | 86.82 | 83.32 | 84.94 | 87.62 | 82.42 |
| BiGRU | 86.00 | 87.50 | 90.45 | 84.88 | 85.92 | 87.47 | 90.12 | 85.10 |
| RCNN | 85.79 | 87.32 | 90.15 | 84.85 | 85.27 | 86.95 | 89.10 | 85.03 |
| CNN | 85.68 | 87.12 | 89.53 | 84.83 | 82.92 | 84.41 | 88.15 | 80.97 |
| CNN-GRU | 85.47 | 87.11 | 89.38 | 85.24 | 85.30 | 86.93 | 89.49 | 84.72 |
| CNN-LSTM | 85.35 | 87.03 | 89.27 | 85.20 | 85.33 | 86.87 | 89.94 | 84.13 |
| GNN | 85.02 | 86.28 | 90.35 | 82.57 | 81.75 | 82.75 | 89.85 | 76.68 |

## 7.4 Performance-Resource Analysis

The experimental results demonstrate distinct performance patterns across different model architectures and computational resources:

*7.4.1 High-Performance Models.* Our analysis reveals several key findings:

- Transformer architectures consistently achieve the highest performance, with RoBERTa leading at 91.89% accuracy and F1-score, followed by BERT (91.20% accuracy and F1-score) on the balanced raw dataset.
- DL approaches also show strong performance, particularly with GloVe embeddings: HAN achieves 87.18% accuracy and 88.64% F1-score, BiGRU reaches 86.70% accuracy and 88.17% F1-score, and BiLSTM attains 86.34% accuracy and 87.88% F1-score on the "Combined Mid." Dataset.
- Transformer models achieved better performance on raw data, with preprocessing steps showing negative impact over unprocessed text input.

*7.4.2 Resource-Efficient Models.* The results demonstrate that:

- Traditional machine learning approaches maintain strong performance while requiring fewer computational resources, with CatBoost achieving 88.60% accuracy and 89.60% F1-score, followed by SVM at 88.50% accuracy and 89.50% F1-score.
- Preprocessing showed no significant benefits for most models. Only SVM demonstrated a minimal improvement (from 88.50% to 88.56%), while other traditional ML models actually performed slightly worse with preprocessed data, as shown in Table 4.
- Balanced datasets of moderate size (267K samples) provide better performance than larger unbalanced datasets (451K samples). Similar results are shown in Table 2 on the Combined Small Dataset which is slightly unbalanced with class distributions 57.87% and 42.13%.

**Table 4: Results of Traditional ML Models with TF-IDF on "Combined Mid." Dataset**

| Model | Accuracy | F1 | Precision | Recall |
|---|---|---|---|---|
| **Raw Data** | | | | |
| CatBoost | **88.60** | **89.60** | 94.90 | 84.80 |
| SVM RBF | 88.50 | 89.50 | 94.30 | 85.20 |
| LightGBM | 88.20 | 89.20 | 94.60 | 84.40 |
| XGBoost | 88.00 | 88.90 | 95.10 | 83.40 |
| Logistic Regression | 87.80 | 89.00 | 92.80 | 85.60 |
| Random Forest | 86.90 | 88.30 | 90.90 | 85.80 |
| AdaBoost | 86.00 | 87.00 | 93.70 | 81.10 |
| Decision Tree | 85.40 | 87.30 | 87.70 | 86.90 |
| Gradient Boosting | 84.90 | 85.40 | **96.70** | 76.40 |
| Multinomial NB | 82.50 | 85.30 | 83.20 | **87.50** |
| Bernoulli NB | 81.60 | 84.10 | 84.00 | 84.20 |
| **Preprocessed Data** | | | | |
| SVM RBF | **88.56** | **89.58** | 94.59 | 85.07 |
| CatBoost | 88.42 | 89.42 | 94.73 | 84.67 |
| LightGBM | 87.95 | 88.97 | 94.42 | 84.11 |
| Logistic Regression | 87.93 | 89.07 | 93.40 | 85.13 |
| Random Forest | 87.79 | 89.14 | 91.71 | 86.71 |
| XGBoost | 87.69 | 88.63 | 95.02 | 83.06 |
| AdaBoost | 86.25 | 87.26 | 93.86 | 81.54 |
| Decision Tree | 85.82 | 87.70 | 87.94 | 87.46 |
| Gradient Boosting | 84.93 | 85.44 | **96.67** | 76.55 |
| Multinomial NB | 82.01 | 85.16 | 81.33 | **89.36** |
| Bernoulli NB | 81.26 | 84.30 | 81.72 | 87.05 |

**Table 5: Unbalanced Preprocessed Dataset Results (451K Samples)**

| Metric | BERT | RoBERTa | XLM | DistilBERT | ALBERT |
|---|---|---|---|---|---|
| Accuracy | 90.71 | **90.72** | 90.41 | 90.36 | 90.11 |
| F1 | **90.57** | 90.54 | 90.31 | 90.25 | 89.92 |
| Precision | **90.49** | 90.46 | 90.26 | 90.19 | 89.85 |
| Recall | 90.71 | **90.72** | 90.41 | 90.36 | 90.11 |

**Table 6: Balanced Raw Dataset Results (267K Samples)**

| Metric | BERT | RoBERTa | XLM | DistilBERT | ALBERT |
|---|---|---|---|---|---|
| Accuracy | 91.20 | **91.89** | 91.08 | 91.10 | 90.46 |
| F1 | 91.20 | **91.89** | 91.07 | 91.10 | 90.45 |
| Precision | 91.28 | **92.01** | 91.20 | 91.22 | 90.57 |
| Recall | 91.20 | **91.89** | 91.08 | 91.10 | 90.46 |

## 8 Conclusion and Future Work

This study presents a systematic evaluation of hate speech detection methods, analyzing 38 distinct model configurations across datasets ranging from 6.5K to 451K samples. Transformer architectures outperform other approaches, with RoBERTa achieving the highest accuracy and F1-score (91.89%) on a balanced raw dataset. Among deep learning models, Hierarchical Attention Networks (HAN) with GloVe embeddings achieve the best results (87.18% accuracy, 88.64% F1-score), followed closely by BiGRU (86.70% accuracy, 88.17% F1-score). Preprocessing notably impacts performance, with transformers showing degradation on preprocessed datasets.

**Table 7: Balanced Preprocessed Dataset Results (160K Samples)**

| Metric | BERT | RoBERTa | XLM | DistilBERT | ALBERT |
|---|---|---|---|---|---|
| Accuracy | 86.32 | **86.46** | 86.04 | 86.19 | 85.73 |
| F1 | 86.32 | **86.45** | 86.04 | 86.18 | 85.71 |
| Precision | 86.33 | **86.61** | 86.10 | 86.27 | 85.88 |
| Recall | 86.32 | **86.46** | 86.04 | 86.19 | 85.73 |

Dataset characteristics significantly influence model performance. The balanced raw dataset (267K samples) consistently outperforms both larger unbalanced datasets (451K samples) and smaller configurations. Traditional machine learning approaches offer an efficient alternative, with CatBoost (88.60% accuracy, 89.60% F1-score) and SVM with RBF kernel (88.50% accuracy, 89.50% F1-score) delivering competitive performance while requiring fewer computational resources.

These findings suggest that for practical implementations, balanced, moderate-sized datasets with raw text provide the most effective foundation for hate speech detection systems, outperforming larger, preprocessed datasets in both accuracy and efficiency.

Several promising directions emerge for future research. These include integrating multi-modal data sources (e.g., text, audio, and images) to enhance detection capabilities beyond text-based analysis, developing architecture-specific preprocessing strategies to optimize performance, exploring optimal dataset scaling with a focus on class distribution balance, and investigating model adaptability to evolving hate speech patterns. Addressing these challenges will help overcome current limitations and contribute to the development of more robust and adaptable detection systems.

## References

[1] Sweta Agrawal and Amit Awekar. 2018. Deep Learning for Detecting Cyberbullying Across Multiple Social Media Platforms. In *European Conference on Information Retrieval*. Springer, Grenoble, France, 141–153.

[2] Fariha Tahosin Boishakhi, Ponkoj Chandra Shill, and Md Golam Rabiul Alam. 2021. Multi-modal Hate Speech Detection Using Machine Learning. In *2021 IEEE International Conference on Big Data (Big Data)*. IEEE, Orlando, Florida, USA, 4496–4499.

[3] Leo Breiman. 2001. Random Forests. *Machine Learning* 45, 1 (2001), 5–32.

[4] Mohit Chandra, Dheeraj Pailla, Himanshu Bhatia, Aadilmehdi Sanchawala, Manish Gupta, Manish Shrivastava, and Ponnurangam Kumaraguru. 2021. Subverting the Jewtocracy: Online Antisemitism Detection Using Multimodal Deep Learning. In *Proceedings of the 13th ACM Web Science Conference 2021*. ACM, Southampton, England, 148–157. https://doi.org/10.1145/3447535.3462502

[5] Tianqi Chen and Carlos Guestrin. 2016. XGBoost: A Scalable Tree Boosting System. In *Proceedings of the 22nd ACM SIGKDD International Conference on Knowledge Discovery and Data Mining*. San Francisco, CA, USA, 785–794.

[6] Kyunghyun Cho, Bart van Merrienboer, Caglar Gulcehre, Dzmitry Bahdanau, Fethi Bougares, Holger Schwenk, and Yoshua Bengio. 2014. Learning Phrase Representations Using RNN Encoder-Decoder for Statistical Machine Translation. In *Proceedings of the 2014 Conference on Empirical Methods in Natural Language Processing (EMNLP)*. Doha, Qatar, 1724–1734.

**Table 8: Transformers Performance Comparison Across All Datasets**

| Dataset | Metric | BERT | RoBERTa | XLM | DistilBERT | ALBERT |
|---|---|---|---|---|---|---|
| Dataset 1 | Accuracy | 85.24 | **85.86** | 84.93 | 85.33 | 84.19 |
| | F1 | 89.26 | **89.67** | 89.05 | 89.35 | 88.42 |
| | Precision | 87.84 | **88.65** | 87.66 | 87.79 | 87.57 |
| | Recall | 90.79 | 90.76 | 90.56 | **90.98** | 89.30 |
| Dataset 2 | Accuracy | 88.57 | **89.29** | 88.34 | 88.45 | 88.42 |
| | F1 | 79.25 | **80.26** | 78.76 | 78.13 | 78.86 |
| | Precision | 76.31 | 78.40 | 76.23 | **78.47** | 77.13 |
| | Recall | **82.64** | 82.50 | 81.71 | 78.13 | 81.49 |
| Dataset 3 | Accuracy | 96.47 | **96.49** | 96.23 | 96.37 | 96.21 |
| | F1 | 97.87 | **97.88** | 97.73 | 97.82 | 97.71 |
| | Precision | 98.14 | **98.39** | 98.18 | 98.03 | 98.10 |
| | Recall | **97.61** | 97.39 | 97.27 | **97.61** | 97.33 |
| Combined | Accuracy | 91.40 | **91.48** | 90.61 | 91.38 | 90.92 |
| | F1 | 92.55 | **92.62** | 91.97 | 92.52 | 92.01 |
| | Precision | 92.68 | 92.73 | 90.84 | 92.72 | **93.64** |
| | Recall | 92.41 | 92.50 | **93.13** | 92.32 | 90.43 |
| LUP (451K) | Accuracy | 90.71 | **90.72** | 90.41 | 90.36 | 90.11 |
| | F1 | **90.57** | 90.54 | 90.31 | 90.25 | 89.92 |
| | Precision | **90.49** | 90.46 | 90.26 | 90.19 | 89.85 |
| | Recall | 90.71 | **90.72** | 90.41 | 90.36 | 90.11 |
| LBR (267K) | Accuracy | 91.20 | **91.89** | 91.08 | 91.10 | 90.46 |
| | F1 | 91.20 | **91.89** | 91.07 | 91.10 | 90.45 |
| | Precision | 91.28 | **92.01** | 91.20 | 91.22 | 90.57 |
| | Recall | 91.20 | **91.89** | 91.08 | 91.10 | 90.46 |
| LBP (160K) | Accuracy | 86.32 | **86.46** | 86.04 | 86.19 | 85.73 |
| | F1 | 86.32 | **86.45** | 86.04 | 86.18 | 85.71 |
| | Precision | 86.33 | **86.61** | 86.10 | 86.27 | 85.88 |
| | Recall | 86.32 | **86.46** | 86.04 | 86.19 | 85.73 |

[7] Alexis Conneau, Kartikay Khandelwal, Naman Goyal, Vishrav Chaudhary, Guillaume Wenzek, Francisco Guzmán, Edouard Grave, Myle Ott, Luke Zettlemoyer, and Veselin Stoyanov. 2020. Unsupervised Cross-lingual Representation Learning at Scale. In *Proceedings of the 58th Annual Meeting of the Association for Computational Linguistics*. Seattle, WA, USA, 8440–8451. https://doi.org/10.18653/v1/2020.acl-main.747

[8] Corinna Cortes and Vladimir Vapnik. 1995. Support-Vector Networks. In *Machine Learning*, Vol. 20. 273–297.

[9] Thomas Davidson, Dana Warmsley, Michael Macy, and Ingmar Weber. 2017. Automated Hate Speech Detection and the Problem of Offensive Language. In *Proceedings of the International AAAI Conference on Web and Social Media*, Vol. 11. AAAI Press, Montreal, Canada, 512–515.

[10] Jacob Devlin, Ming-Wei Chang, Kenton Lee, and Kristina Toutanova. 2019. BERT: Pre-training of Deep Bidirectional Transformers for Language Understanding. In *Proceedings of the 2019 Conference of the North American Chapter of the Association for Computational Linguistics: Human Language Technologies, Volume 1 (Long and Short Papers)*. Association for Computational Linguistics, Minneapolis, MN, USA, 4171–4186. https://doi.org/10.18653/v1/N19-1423

[11] Mai ElSherief, Vivek Kulkarni, Dana Nguyen, William Yang Wang, and Elizabeth Belding. 2018. Hate Lingo: A Target-based Linguistic Analysis of Hate Speech in Social Media. In *Proceedings of the International AAAI Conference on Web and Social Media*, Vol. 12. Palo Alto, California, USA, 321–330.

[12] Paula Fortuna and Sérgio Nunes. 2019. A Survey on Automatic Detection of Hate Speech in Text. *Comput. Surveys* 51, 4 (2019), 1–30. https://doi.org/10.1145/3232676

[13] Yoav Freund and Robert E. Schapire. 1997. A Decision-Theoretic Generalization of On-Line Learning and an Application to Boosting. *J. Comput. System Sci.* 55, 1 (1997), 119–139.

[14] Jerome H. Friedman. 2001. Greedy Function Approximation: A Gradient Boosting Machine. *Annals of Statistics* (2001), 1189–1232.

[15] Kyle Gallatin and Chris Albon. 2023. *Machine Learning with Python Cookbook: Practical Solutions from Preprocessing to Deep Learning*. O'Reilly Media, Boston, MA, USA.

[16] Gary W Giumetti and Robin M Kowalski. 2022. Cyberbullying via Social Media and Well-being. *Current Opinion in Psychology* 45 (2022), 101314. https://doi.org/10.1016/j.copsyc.2022.101314

[17] Alex Graves and Jürgen Schmidhuber. 2005. Framewise Phoneme Classification with Bidirectional LSTM and Other Neural Network Architectures. *Neural Networks* 18, 5–6 (2005), 602–610.

[18] Khalid Hasan and Jamil Saquer. 2024. A Comparative Analysis of Transformer and LSTM Models for Detecting Suicidal Ideation on Reddit. In *2024 International Conference on Machine Learning and Applications (ICMLA)*. Miami, FL, USA, 1343 – 1349. https://doi.org/10.1109/ICMLA61862.2024.00209

[19] Sepp Hochreiter and Jürgen Schmidhuber. 1997. Long Short-Term Memory. *Neural Computation* 9, 8 (1997), 1735–1780.

[20] David W. Jr Hosmer, Stanley Lemeshow, and Rodney X. Sturdivant. 2013. *Applied Logistic Regression*. John Wiley & Sons.

[21] Impermium. 2012. Detecting Insults in Social Commentary Dataset. Retrieved from https://www.kaggle.com/c/detecting-insults-in-social-commentary.

[22] Guolin Ke, Qi Meng, Thomas Finley, Taifeng Wang, Wei Chen, Weidong Ma, Qiwei Ye, and Tie-Yan Liu. 2017. LightGBM: A Highly Efficient Gradient Boosting Decision Tree. *Advances in Neural Information Processing Systems* 30 (2017), 3146–3154.

[23] Heena Khan and Joshua L. Phillips. 2021. Language Agnostic Model: Detecting Islamophobic Content on Social Media. In *Proceedings of the 2021 ACM Southeast Conference*. ACM, Jacksonville, AL, USA, 229–233.

[24] Yoon Kim. 2014. Convolutional Neural Networks for Sentence Classification. In *Proceedings of the 2014 Conference on Empirical Methods in Natural Language Processing (EMNLP)*. Doha, Qatar, 1746–1751.

[25] Siwei Lai, Liheng Xu, Kang Liu, and Jun Zhao. 2015. Recurrent Convolutional Neural Networks for Text Classification. In *Proceedings of the Twenty-ninth AAAI Conference on Artificial Intelligence*. Austin, TX, USA.

[26] Zhenzhong Lan, Mingda Chen, Sebastian Goodman, Kevin Gimpel, Piyush Sharma, and Radu Soricut. 2020. ALBERT: A Lite BERT for Self-supervised Learning of Language Representations. arXiv:1909.11942 https://arxiv.org/abs/1909.11942

[27] Yann LeCun, Léon Bottou, Yoshua Bengio, and Patrick Haffner. 1998. Gradient-based Learning Applied to Document Recognition. *Proc. IEEE* 86, 11 (1998), 2278–2324.

[28] Richard Liaw, Eric Liang, Robert Nishihara, Philipp Moritz, Joseph E Gonzalez, and Ion Stoica. 2018. Tune: A Research Platform for Distributed Model Selection and Training. *arXiv preprint arXiv:1807.05118* (2018).

[29] Anna Liu. 2018. *Neural Network Models for Hate Speech Classification in Tweets*. Ph. D. Dissertation. Harvard University, Cambridge, MA, USA. https://dash.harvard.edu/handle/1/38811552 Ph.D. Dissertation.

[30] Yinhan Liu, Myle Ott, Naman Goyal, Jingfei Du, Mandar Joshi, Danqi Chen, Omer Levy, Mike Lewis, Luke Zettlemoyer, and Veselin Stoyanov. 2019. RoBERTa: A Robustly Optimized BERT Pretraining Approach. In *International Conference on Learning Representations*. OpenReview, New Orleans, LA, USA, 7550–7564.

[31] Sean MacAvaney, Hao-Ren Yao, Eugene Yang, Katina Russell, Nazli Goharian, and Ophir Frieder. 2019. Hate Speech Detection: Challenges and Solutions. *PloS One* 14, 8 (2019).

[32] Andrew McCallum and Kamal Nigam. 1998. A Comparison of Event Models for Naive Bayes Text Classification. In *AAAI-98 Workshop on Learning for Text Categorization*. Madison, WI, USA, 41–48.

[33] Tomas Mikolov, Kai Chen, Greg Corrado, and Jeffrey Dean. 2013. Efficient Estimation of Word Representations in Vector Space. *arXiv preprint arXiv:1301.3781* 3781 (2013).

[34] Devansh Mody, YiDong Huang, and Thiago Eustaquio Alves de Oliveira. 2023. A Curated Dataset for Hate Speech Detection on Social Media Text. *Data in Brief* 46 (2023), 108832.

[35] Jeffrey Pennington, Richard Socher, and Christopher D Manning. 2014. Glove: Global Vectors for Word Representation. In *Proceedings of the 2014 Conference on Empirical Methods in Natural Language Processing (EMNLP)*. Doha, Qatar, 1532–1543.

[36] Liudmila Prokhorenkova, Gleb Gusev, Aleksandr Vorobev, Anna Veronika Dorogush, and Andrey Gulin. 2018. CatBoost: Unbiased Boosting with Categorical Features. In *Advances in Neural Information Processing Systems*. Montréal, QC, Canada, 6638–6648.

[37] J. Ross Quinlan. 1986. Induction of Decision Trees. *Machine Learning* 1, 1 (1986), 81–106.

[38] Björn Ross, Michael Rist, Guillermo Carbonell, Benjamin Cabrera, Nils Kurowsky, and Michael Wojatzki. 2017. Measuring the Reliability of Hate Speech Annotations: The Case of the European Refugee Crisis. *arXiv preprint arXiv:1701.08118* (2017).

[39] Hind Saleh, Areej Alhothali, and Kawthar Moria. 2023. Detection of Hate Speech Using BERT and Hate Speech Word Embedding with Deep Model. *Applied Artificial Intelligence* 37, 1 (2023). https://doi.org/10.1080/08839514.2023.2166719

[40] Calvin Erico Rudy Salim and Derwin Suhartono. 2020. A Systematic Literature Review of Different Machine Learning Methods on Hate Speech Detection. *JOIV: International Journal on Informatics Visualization* 4, 4 (2020), 213–218.

[41] Victor Sanh, Lysandre Debut, Julien Chaumond, and Thomas Wolf. 2020. DistilBERT, A Distilled Version of BERT: Smaller, Faster, Cheaper and Lighter. arXiv:1910.01108 https://arxiv.org/abs/1910.01108

[42] Anna Schmidt and Michael Wiegand. 2017. A Survey on Hate Speech Detection Using Natural Language Processing. In *Proceedings of the Fifth International Workshop on Natural Language Processing for Social Media*. Association for Computational Linguistics, Valencia, Spain, 1–10.

[43] Nabil Shawkat, Jamil Saquer, and Hazim Shatnawi. 2024. Evaluation of Different Machine Learning and Deep Learning Techniques for Hate Speech Detection. In *2024 ACM Southeast Conference (ACMSE 2024)*. ACM, Marietta, GA, USA, 1–6. https://doi.org/10.1145/3603287.3651218

[44] Nabil Shawkat, Jesse Simpson, and Jamil Saquer. 2022. Evaluation of Different ML and Text Processing Techniques for Hate Speech Detection. In *IEEE 4th International Conference on Data Intelligence and Security (ICDIS 2022)*. IEEE, Shenzhen, China, 213–219.

[45] Marco Siino, Ilenia Tinnirello, and Marco La Cascia. 2024. Is Text Preprocessing Still Worth the Time? A Comparative Survey on the Influence of Popular Preprocessing Methods on Transformers and Traditional Classifiers. *Information Systems* 121 (2024), 102342.

[46] Mifta Sintaha and Moin Mostakim. 2018. An Empirical Study and Analysis of the Machine Learning Algorithms Used in Detecting Cyberbullying in Social Media. In *2018 21st International Conference on Computer and Information Technology (ICCIT)*. IEEE, Dhaka, Bangladesh, 1–6. https://doi.org/10.1109/ICCITECHN.2018.8631958

[47] Duyu Tang, Bing Qin, and Ting Liu. 2015. Document Modeling with Gated Recurrent Neural Network for Sentiment Classification. In *Proceedings of the 2015 Conference on Empirical Methods in Natural Language Processing*. Lisbon, Portugal, 1422–1432.

[48] Zeerak Waseem and Dirk Hovy. 2016. Hateful Symbols or Hateful People? Predictive Features for Hate Speech Detection on Twitter. In *Proceedings of the NAACL Student Research Workshop*. Association for Computational Linguistics, San Diego, CA, USA, 88–93.

[49] Michael Wiegand, Josef Ruppenhofer, and Thomas Kleinbauer. 2019. Detection of Abusive Language: The Problem of Biased Datasets. In *Proceedings of the 2019 Conference of the North American Chapter of the Association for Computational Linguistics: Human Language Technologies, Volume 1 (Long and Short Papers)*. Minneapolis, Minnesota, USA, 602–608.

[50] Zonghan Wu, Shirui Pan, Fengwen Chen, Guodong Long, Chengqi Zhang, and Philip S Yu. 2020. A Comprehensive Survey on Graph Neural Networks. *IEEE Transactions on Neural Networks and Learning Systems* 32, 1 (2020), 4–24.

[51] Ellery Wulczyn, Nithum Thain, and Lucas Dixon. 2017. Ex Machina: Personal Attacks Seen at Scale. In *Proceedings of the 26th International Conference on World Wide Web*. Perth, Australia, 1391–1399. https://doi.org/10.1145/3038912.3052591

[52] Zichao Yang, Diyi Yang, Chris Dyer, Xiaodong He, Alexander J Smola, and Eduard H Hovy. 2016. Hierarchical Attention Networks for Document Classification. In *Proceedings of the 2016 Conference of the North American Chapter of the Association for Computational Linguistics: Human Language Technologies*. San Diego, CA, USA, 1480–1489.

[53] Chunting Zhou, Chonglin Sun, Zhiyuan Liu, and Francis Lau. 2015. A C-LSTM Neural Network for Text Classification. *arXiv preprint arXiv:1511.08630* (2015).

## Appendix A - Additional Results

**Table A.1: Results of Traditional Deep Learning Models on Small and Large Datasets (%)**

| | Dataset1 | | Dataset2 | | Dataset3 | | Dataset5 (160K) | | Dataset6 (267K) | | Dataset7 (451K) | |
|---|---|---|---|---|---|---|---|---|---|---|---|---|
| **Model** | **Acc** | **F1** | **Acc** | **F1** | **Acc** | **F1** | **Acc** | **F1** | **Acc** | **F1** | **Acc** | **F1** |
| HAN_GloVe | **81.83** | **87.05** | 82.86 | 64.99 | **94.90** | **96.92** | 82.64 | 83.28 | 85.42 | 85.73 | 87.78 | 87.63 |
| BiGRU_GloVe | 81.69 | 86.96 | 81.97 | 64.43 | 94.38 | 96.60 | 82.30 | 82.95 | 85.50 | 85.80 | 87.82 | 87.35 |
| BiLSTM_GloVe | 81.65 | 86.73 | 81.67 | 63.17 | 94.34 | 96.59 | 82.37 | 82.83 | 85.51 | 85.78 | 87.95 | 87.49 |
| HAN_Word2Vec | 81.63 | 86.89 | **83.99** | **68.12** | 94.83 | 96.87 | 82.74 | 83.41 | 85.58 | 85.98 | 87.85 | **87.71** |
| BiLSTM_Word2Vec | 81.13 | 86.58 | 83.62 | 66.22 | 94.00 | 96.37 | 82.86 | 83.53 | 85.62 | 86.02 | **88.02** | 87.67 |
| BiGRU_Word2Vec | 81.00 | 86.60 | 83.06 | 67.05 | 94.56 | 96.72 | **82.90** | **83.68** | 85.37 | 85.69 | 87.74 | 87.39 |
| LSTM_GloVe | 80.87 | 86.55 | 79.83 | 60.88 | 92.35 | 95.39 | 80.66 | 81.26 | 83.59 | 83.65 | 86.42 | 84.99 |
| GRU_GloVe | 80.78 | 86.47 | 81.80 | 66.20 | 92.50 | 95.45 | 80.55 | 81.23 | 83.43 | 83.69 | 86.54 | 85.06 |
| GRU_Word2Vec | 80.78 | 86.56 | 82.03 | 63.48 | 92.05 | 95.18 | 80.64 | 81.54 | 83.19 | 83.31 | 86.76 | 85.51 |
| TextCNN_Word2Vec | 80.65 | 86.52 | 83.53 | 65.93 | 94.24 | 96.52 | 82.36 | 83.15 | **85.66** | **86.04** | 87.37 | 87.39 |
| RCNN_Word2Vec | 80.42 | 86.16 | 83.41 | 65.62 | 94.40 | 96.63 | 82.12 | 82.88 | 85.32 | 85.65 | 87.57 | 87.33 |
| CNN-GRU_Word2Vec | 80.33 | 86.26 | 83.65 | 65.52 | 94.15 | 96.47 | 81.89 | 82.72 | 85.14 | 85.40 | 87.38 | 87.15 |
| CNN-LSTM_Word2Vec | 80.15 | 86.10 | 82.82 | 66.12 | 94.17 | 96.49 | 81.99 | 82.78 | 85.23 | 85.48 | 87.30 | 87.05 |
| TextCNN_GloVe | 80.14 | 86.24 | 81.38 | 62.64 | 93.68 | 96.17 | 81.19 | 82.15 | 84.95 | 85.42 | 86.72 | 86.46 |
| LSTM_Word2Vec | 80.09 | 86.06 | 83.40 | 62.94 | 91.93 | 95.13 | 81.01 | 81.67 | 83.27 | 83.16 | 86.88 | 85.55 |
| CNN_Word2Vec | 80.03 | 86.38 | 82.26 | 64.55 | 92.07 | 95.23 | 79.49 | 80.57 | 83.70 | 84.26 | 82.27 | 74.42 |
| CNN-GRU_GloVe | 79.67 | 85.85 | 80.60 | 60.29 | 93.52 | 96.07 | 81.21 | 82.03 | 84.55 | 84.96 | 87.00 | 86.64 |
| CNN_GloVe | 79.49 | 85.80 | 81.27 | 61.22 | 91.29 | 94.70 | 78.84 | 80.22 | 82.60 | 82.89 | 85.51 | 83.18 |
| RCNN_GloVe | 79.47 | 85.59 | 81.44 | 62.29 | 93.47 | 96.03 | 81.08 | 81.97 | 84.64 | 84.93 | 87.29 | 86.73 |
| GNN_GloVe | 79.19 | 85.41 | 77.63 | 50.08 | 90.56 | 94.41 | 77.31 | 78.50 | 80.05 | 80.89 | 82.15 | 74.85 |
| CNN-LSTM_GloVe | 78.96 | 85.23 | 80.85 | 58.36 | 93.14 | 95.86 | 80.97 | 81.80 | 84.51 | 84.80 | 87.14 | 86.85 |
| GNN_Word2Vec | 78.44 | 84.99 | 80.14 | 55.14 | 90.66 | 94.44 | 78.13 | 79.23 | 80.81 | 81.42 | 86.04 | 83.98 |

**Table A.2: Results of Traditional Machine Learning Models on Small and Large Datasets (%)**

| | Dataset1 | | Dataset2 | | Dataset3 | | Dataset5 (160K) | | Dataset6 (267K) | | Dataset7 (451K) | |
|---|---|---|---|---|---|---|---|---|---|---|---|---|
| **Model** | **Acc** | **F1** | **Acc** | **F1** | **Acc** | **F1** | **Acc** | **F1** | **Acc** | **F1** | **Acc** | **F1** |
| RandomForest_TFIDF | **83.81** | **88.70** | 82.71 | 60.82 | 95.02 | 97.00 | 81.36 | 81.46 | 88.46 | 88.53 | 87.97 | 87.11 |
| SVM_RBF_TFIDF | 83.03 | 88.35 | 82.11 | 56.13 | 95.38 | 97.21 | **83.50** | **83.66** | **88.71** | **88.73** | **88.30** | **87.38** |
| LogisticRegression_TFIDF | 82.40 | 87.95 | 82.41 | 56.72 | 94.63 | 96.80 | 81.69 | 81.75 | 83.94 | 83.69 | 86.96 | 85.83 |
| BernoulliNB_TFIDF | 82.16 | 87.48 | 82.34 | 61.23 | 91.51 | 95.02 | 68.82 | 74.44 | 68.79 | 74.06 | 57.54 | 62.11 |
| XGBoost_TFIDF | 81.89 | 87.51 | 81.73 | 59.22 | 95.82 | 97.44 | 79.40 | 78.52 | 80.66 | 79.19 | 86.38 | 84.16 |
| LightGBM_TFIDF | 81.83 | 87.33 | 82.11 | 60.67 | 95.76 | 97.41 | 79.64 | 78.94 | 80.62 | 79.21 | 86.21 | 84.19 |
| CatBoost_TFIDF | 81.65 | 87.43 | **83.09** | 60.67 | **96.11** | **97.62** | 81.71 | 81.42 | 83.94 | 83.39 | 87.29 | 85.88 |
| AdaBoost_TFIDF | 81.26 | 87.28 | 82.71 | **61.87** | 95.44 | 97.19 | 69.93 | 63.00 | 71.12 | 63.54 | 82.33 | 76.52 |
| GradientBoosting_TFIDF | 80.78 | 87.06 | 81.88 | 54.99 | 91.75 | 95.15 | 71.28 | 75.32 | 74.31 | 69.80 | 84.00 | 78.96 |
| MultinomialNB_TFIDF | 78.95 | 86.04 | 77.48 | 31.72 | 86.46 | 92.39 | 79.34 | 80.54 | 81.19 | 81.64 | 85.77 | 84.54 |
| DecisionTree_TFIDF | 78.20 | 84.04 | 78.09 | 59.35 | 94.75 | 96.82 | 77.02 | 77.53 | 84.86 | 85.11 | 85.40 | 85.25 |